\documentclass[11pt,letterpaper]{article}
\usepackage[letterpaper]{geometry}
\usepackage{acl2012}
\usepackage{microtype} %Magic, make the paper more dense!

\makeatletter
\newcommand{\@BIBLABEL}{\@emptybiblabel}
\newcommand{\@emptybiblabel}[1]{}
\makeatother

\usepackage{times}
\usepackage{latexsym}

\usepackage{url}
\usepackage{placeins}
\usepackage{longtable}
\usepackage{pgfplots}

\pgfmathdeclarefunction{gauss}{2}{%
  \pgfmathparse{1/(#2*sqrt(2*pi))*exp(-((x-#1)^2)/(2*#2^2))}%
}

\pgfmathdeclarefunction{gaussBias}{3}{%
  \pgfmathparse{#3+1/(#2*sqrt(2*pi))*exp(-((x-#1)^2)/(2*#2^2))}%
}

\usepackage[colorlinks=true,citecolor=blue,urlcolor=blue,linkcolor=blue,pagebackref=false,hypertexnames=false]{hyperref}

\hypersetup{citecolor=[rgb]{0.078,0.337,0.5},urlcolor=[rgb]{0.078,0.337,0.5},linkcolor=[rgb]{0.078,0.337,0.5}}

\usepackage{graphicx}
\usepackage{multirow}

\usepackage{amsmath}

\usepackage{amssymb}
\usepackage[parfill]{parskip}
 \usepackage{array}
\newcolumntype{P}[1]{>{\centering\arraybackslash}p{#1}}
  
\newcommand{\refsec}[1]{\hyperref[#1]{section \ref*{#1}}}
\newcommand{\refsecUpper}[1]{\hyperref[#1]{\mbox{Section \ref*{#1}}}}
\newcommand{\reftable}[1]{\hyperref[#1]{\mbox{Table \ref*{#1}}}}
\newcommand{\reffigure}[1]{\hyperref[#1]{\mbox{Figure \ref*{#1}}}}

\title{Why Comparing Single Performance Scores Does Not Allow to Draw Conclusions About Machine Learning Approaches}

\author{Nils Reimers \and Iryna Gurevych \\
Ubiquitous Knowledge Processing Lab (UKP) and Research Training Group AIPHES\\
Department of Computer Science, Technische Universit\"at Darmstadt\\
\url{www.ukp.tu-darmstadt.de}}
\date{}

\begin{document}
\maketitle
\begin{abstract}
Developing \textit{state-of-the-art approaches} for specific tasks is a major driving force in our research community. Depending on the prestige of the task, publishing it can come along with a lot of visibility. The question arises how reliable are our evaluation methodologies to compare approaches? 

One common methodology to identify the state-of-the-art is to partition data into a train, a development and a test set. Researchers can train and tune their approach on some part of the dataset and then select the model that worked best on the development set for a final evaluation on unseen test data. Test scores from different approaches are  compared, and performance differences are tested for statistical significance.

In this publication, we show that there is a high risk that a statistical significance in this type of evaluation is not due to a superior learning approach. Instead, there is a high risk that the difference is due to chance. For example for the CoNLL 2003 NER dataset we observed in up to 26\% of the cases type I errors (false positives) with a threshold of $p < 0.05$, i.e., falsely concluding a statistically significant difference between two identical approaches.

We prove that this evaluation setup is unsuitable to compare learning approaches. We formalize alternative evaluation setups based on score distributions. 
\end{abstract}

\section{Introduction} \label{intro}
Given two machine learning approaches, approach \textbf{A} and approach \textbf{B}, for a certain dataset, how can we decide which approach is more accurate for this task? This is a fundamental question in our community, where a lot of effort is spent to identify new \textit{state-of-the-art} approaches. Hence, we want that the evaluation setup is not impacted by random chance and we should draw the same conclusion if the experiment is reproduced.

While different evaluation setups exist, one fairly common evaluation setup is to partition annotated data into a training, development and test set. Approaches are trained and tuned on the train and development set, and then a performance score on a held-out test set is computed. The approach with the higher test performance score is observed as superior\footnote{In this paper, we only judge approaches based on how accurate those are given a specific performance measure. For real-world applications, superiority can mean many distinct things that are not related to accuracy. }. 

As the test set is a finite sample, the test score differs from the (hypothetical) performance on the complete data distribution. A significance test on the test set is used to reduce the risk that chance induced from the finite test sample is the explanation for the difference. If the difference is significant, it is usually accepted that one approach is superior to the other. 

\begin{figure}[ht]
 \centering
  \includegraphics[width=.35\textwidth]{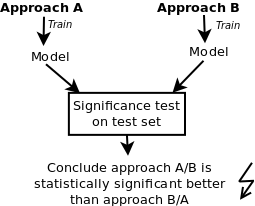}
 \caption{\small Common evaluation methodology to compare two approaches for a specific task.}
\label{fig_prob_sig_test_wrong}
\end{figure}

This evaluation methodology is often used in scientific publications and for shared tasks in our field, for example, it is commonly used for the shared tasks at the International Workshop on Semantic Evaluation (SemEval) and for the shared tasks from the Conference on Computational Natural Language Learning (CoNLL). The participants either submit the output of their system for the unlabeled test data to the task organizers or, as it was the case for the  CoNLL 2017 shared task on multilingual parsing  \mbox{\cite{Zeman2017}}, participants submitted their system to a cloud-based evaluation platform where it was applied to new data. To identify if differences are significant, the organizers used paired bootstrap resampling. Hiding the test data from the participants eliminates the risk that information about the test data is used for the design of the approach. Depending on the prestige of the shared task, winning it can come along with a lot of visibility. The winning approach is often part of future research or serves as a baseline for new approaches.

The question arises how reliable is this evaluation setup and how reliable are shared tasks to identify the best approach? As our results show, this evaluation methodology is incapable to distinguish which learning approach is superior for the studied task.

In this paper, we show that there is a high risk that chance, and not a superior design, leads to significant differences. For example, for the CoNLL 2003 shared task on NER, we compared two identical neural networks with each other. In 22\% of the cases, we observed a significant difference in test score with $p < 0.05$. By implication, if we observe a significant difference in test performance, we cannot be certain if the difference is due to a superior approach or due to luck. The issue is not a flawed significance test but lies in wrongly drawn conclusions.

In the context of this paper, it is important to notice the difference between \textit{models} and \textit{learning approach}. A \textit{learning approach} describes the holistic setup to solve a certain optimization problem. For neural networks, this would be the network architecture, the optimization algorithm, the loss-function etc.  A \textit{model} is a specific configuration of the weights for this architecture. 

A significance test for a specific model can only check if the model will likely perform better for the whole data distribution. However, we often observe that the conclusion is drawn that a superior model implies a superior  approach for that task. For example, for the shared task SemEval-2017 on semantic textual similarity (STS) the task organizers conclude that the model from the winning team is \textit{``the best overall system''} \cite{Cer2017}. \newcite{Szegedy2014} conclude that the winning model from Clarifai for the ImageNet 2013 challenge was the \textit{``year's best approach''}.

The contribution in this paper is to show, that this conclusion cannot be drawn for \textit{non-deterministic learning approaches}\footnote{We define a learning approach as non-deterministic if it uses a sequence of random numbers to solve the optimization problem. Our observations are extendable to deterministic approaches that have tunable hyperparameters.}, like neural networks. Generating a model with superior (test) performance does not allow the conclusion that the learning approach is superior for that task and data split. If two similar approaches are compared, then there is a high risk that a luckier sequence of random numbers, and not the architecture, decides which approach generates a significantly better test performance.

We argue for a change in the evaluation paradigm of machine learning systems. Instead of comparing and reporting individual system runs, we propose training approaches multiple times and comparing score distributions (\refsec{sec_definition_score_distributions}).

\section{Related Work} \label{sec_related_work}
No evaluation setup is perfect and many points are discussable, for example, the right evaluation metric, how to aggregate results, and many more points \cite{Japkowicz_2011}. With a different evaluation setup, we might draw different conclusions. However, to allow a comparison of approaches, the community often uses common evaluation setups. In a lot of cases, these evaluation setups were established in shared tasks and are used long after the shared task. For example, the dataset and the setup of the CoNLL 2003 shared task on NER are still widely used to evaluate new approaches to detect named entities.

One commonly used methodology to compare machine learning approaches is described by \newcite{Bishop2006} (p. 32): ``\textit{If data is plentiful, then one approach is simply to use some of the available data to train a range of models, [...], and then to compare them on independent data, sometimes called a \textit{validation set}, and select the one having the best predictive performance. [...] it may be necessary to keep aside a third \textit{test set} on which the performance of the selected model is finally evaluated.}''

In order to make contributions by different researchers comparable, a popular tool is to use common dataset. Well known examples are the CoNLL 2003 dataset for NER or the CoNLL 2009 dataset for parsing. For those tasks and datasets, new approaches are trained on the provided data, and the test score is compared against published results.

As the test set is finite in size, there is a chance that a model achieves a better score on the test set, but would not yield a better score on the data population as a whole. To guard against this case, a significance test like the approximate randomized test \cite{Riezler2005} or the bootstrap test \cite{Berg-Kirkpatrick2012} can be applied. Those methods test the null hypothesis that both models would perform equally on the population as a whole. Significance tests typically estimate the confidence $p$, which should be an upper-bound for the probability of a type I error (a false positive error).

Training non-deterministic approaches a single time and comparing test scores can be misleading. It is known that, for example, neural networks converge to different points depending on the sequence of random numbers. However, not all convergence points generalize equally well to unseen data \cite{Hochreiter1997_Flat_Minima,LeCun1998,Erhan2010}. In our previous publication \cite{Reimers2017EMNLP}, we showed for the BiLSTM-CRF architecture for NLP sequence tagging tasks that the performance can vary depending on the random seed value. For the system by \newcite{Ma2016} we showed that the $F_1$-score on the CoNLL 2003 NER dataset can vary between $89.88\%$ and $91.00\%$ and for the system by \newcite{Lample2016} that the performance can vary between $90.19\%$ and $90.81\%$ depending on the random seed value. For some random seed values, the network converged to a poor minimum that generalizes badly on unseen data.

However, we are often only interested in the best performance an approach can achieve, for example, after tuning the approach. Failed attempts, like a random initialization that converged to a poor minimum, are often ignored. We eliminate these failed attempts by evaluating the models on a development set. For the final evaluation, we select only the model that performed best on the development set. The question arises if this is a valid evaluation methodology to compare learning approaches for a task?

To our knowledge, this has not been studied before. In \refsec{sec_best_run_comparison} we formalize this type of evaluation. In \refsec{sec_1_out_of_n} we show empirically for seven NLP sequence tagging tasks that this evaluation method is incapable to compare learning approaches. We then present a proof in \refsec{sec_proof} that this evaluation method is in general incapable to compare learning approaches for any tasks, learning approach, and statistical significance test.

\section{Evaluation Methodologies based on Single Scores} \label{sec_definition_single_score}

This section formalizes evaluation methods that are based on single model comparisons. Note, in all cases we assume a fixed train, development, and test set for example from a shared task.

\subsection{Single Run Comparison}
The first evaluation method is to train both approaches a single time and to compare the test scores.

\textbf{Evaluation 1.} Given two approaches, we train both approaches a single time to generate the models $A_i$ and $B_j$.  We define $\Psi^{(Test)}_{A_i}$ as the test score for model $A_i$ and $\Psi^{(Test)}_{B_j}$ as the test score for model $B_j$. We call approach $A$ is superior over approach $B$ if and only if $\Psi^{(test)}_{A_1} > \Psi^{(test)}_{B_1}$ and the difference is  statistical significant. Commonly used significance tests are an approximate randomized test or a bootstrap test \mbox{\cite{Riezler2005}}.

\begin{figure}[h]
 \centering
  \includegraphics[width=.40\textwidth]{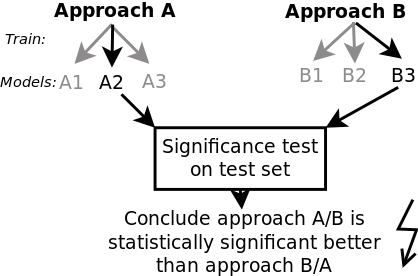}
 \caption{\small Single score comparison for non-deterministic learning approaches (Evaluation 1).}
\label{fig_prob_sig_test_correct}
\end{figure}

Non-deterministic learning approaches, like neural networks, can produce many distinct models $A_1, ..., A_n$. Which model will be produced depends on the sequence of random numbers and cannot be determined in advance.

\reffigure{fig_prob_sig_test_correct} illustrates the issue of this evaluation methodology for non-deterministic learning approaches. Approach $A$ produces the model $A_2$, while approach $B$ the model $B_3$. Model $A_2$ might be significantly better than $B_3$, however, it might be worse than the other models $B_1$ or $B_2$.

\subsection{Best Run Comparison} \label{sec_best_run_comparison}
\begin{figure}[h]
 \centering
  \includegraphics[width=.35\textwidth]{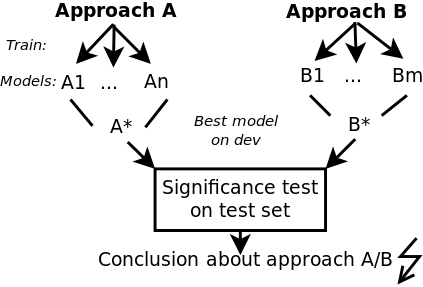}
 \caption{\small Illustration of model tuning and comparing the best models $A_*$ and $B_*$ (Evaluation 2).}
\label{fig_prob_sig_test_def2}
\end{figure}

For shared tasks, the participants are not restricted to train their approach only once. Instead, they can train multiple models and can tune the parameters on the development set. For the final evaluation, they usually must select one model that is compared to the submissions from other participants. A similar process can often be found in scientific publications, where authors tune the approach on a development set and report the test score from the model that performed best on the development set. This form of evaluation is formalized in the following (depicted in \reffigure{fig_prob_sig_test_def2}).

\textbf{Evaluation 2.} Given two approaches and we sample from each multiple models. Approach $A$ produces the models $A_1, ..., A_n$ and approach $B$ the models $B_1, ..., B_m$ with sufficiently large numbers of $n$ and $m$. We define $A_*$ as the best model from approach $A$ and $B_*$ as the best model from approach $B$. \newcite{Bishop2006} defines the best model as the model that performed best on the unseen development set:
\begin{align*}
  A_* &= \text{argmax}_{A_{i} \in \{A_1, ... A_n\}}( \Psi^{(dev)}_{A_{i}}) \\
  B_* &= \text{argmax}_{B_{i} \in \{B_1, ... B_m\}}( \Psi^{(dev)}_{B_{i}})
\end{align*}

With $ \Psi^{(dev)}$ the performance score on the development set. We call approach $A$ is superior over approach $B$ iff $\Psi^{(test)}_{A_*} > \Psi^{(test)}_{B_*}$ and the difference is  significant.

The main contribution in this paper is to show that the conclusion $\Psi^{(test)}_{A_*} > \Psi^{(test)}_{B_*} \Rightarrow$ \textit{approach $A$ better than approach $B$ } is wrong. This implies that this evaluation methodology is unsuitable for shared tasks and research publications.

\section{Experimental Setup}\label{sec_experiment}
We demonstrate that Evaluation 1 and Evaluation 2 fail to identify that two learning approaches are the same. By implication, a significant difference in test score does not allow the conclusion that one approach is better than the other. 

We compare a learning approach $A$ against itself, which we call approach $A$ and $\tilde A$ hereafter.  Approach $A$ and $\tilde A$ use the same code, with the same configuration and are executed on the same computer. The only difference is that the sequence of random number changes each time. 

A suitable evaluation method should conclude that there is no significant difference between $A$ and $\tilde A$ in most cases. We use $p=0.05$ as a threshold, hence, we would expect that a significant difference between $A$ and $\tilde A$ only occurs in at most $5\%$ of the cases.

\subsection{Datasets}
As benchmark tasks, we use seven common NLP sequence tagging tasks. We use the CoNLL 2000 dataset for Chunking, the CoNLL 2003 NER dataset for Named Entity Recognition for English and for German, the ACE 2005 dataset with  the split by \mbox{\newcite{Li_2013}} for entity and event detection, the TempEval 3 event detection dataset\footnote{We used a random fraction of the documents in the training set to form a development set with approximately the size of the test set.}, and the GermEval 2014 dataset for NER in German. We evaluate all tasks in terms of $F_1$-score.

\subsection{Network Architecture}
We use the BiLSTM-CRF architecture we described in \cite{Reimers_2017_Hyperparameter}.\footnote{\url{https://github.com/UKPLab/emnlp2017-bilstm-cnn-crf}} We use 2 hidden layers, 100 hidden units each, variational dropout \mbox{\cite{Gal2015}} of 0.25 applied to both dimensions, Nadam as optimizer \cite{Nadam}, and a mini-batch size of 32 sentences. For the English datasets, we use the pre-trained embeddings by \mbox{\newcite{Komninos2016}}. For the German datasets we used the embeddings by \mbox{\newcite{Reimers2014}}.

\subsection{Training}
In total, we trained 100,000 models for each task with different random seed values. We randomly assign 50,000 models to approach $A$ while the other models are assigned to approach $\tilde A$.

For simplification, we write those models as two matrices with 50 columns and 1,000 rows each:
\begin{align*}
[A_i^{(j)}] \ \ \ \ \ \ \ \ \ \ [\tilde A_i^{(j)}]
\end{align*}
with $i=1,\dots,50$ and $j=1,\dots,1000$. Each model $A_i^{(j)}$ has a development score $\Psi^{(dev)}_{A_i^{(j)}}$ and test score $\Psi^{(test)}_{A_i^{(j)}}$.

Model $A_*^{(j)}$ marks the model with the highest development score from the row $A_{1\leq i \leq 50}^{(j)}$ and $\tilde A_*^{(j)}$ is the model with the highest development score from $\tilde A_{1\leq i \leq 50}^{(j)}$. Hence, we test Evaluation 2 with $n=m=50$.

\subsection{Statistical Significance Test}
We use the bootstrap method by \newcite{Berg-Kirkpatrick2012} with 10,000 samples to test for statistical significance between test performances with a threshold of $p < 0.05$. We also tested the approximate randomized test, and the results were similar.

For Evaluation 1, we test on statistical significance between the models $A_i^{(j)}$ and $\tilde A_i^{(j)}$ for all $i$ and $j$. For Evaluation 2, we test on statistical significance between  $A_*^{(j)}$ and $\tilde A_*^{(j)}$ for $j=1,\dots,1000$.

\section{Results}\label{sec_results_single_score}

\begin{table*}[t]
%\small
\centering
\begin{tabular}{|c|c|c||c|c|}
\hline
\textbf{Task} & \textbf{Threshold $\tau$} & \textbf{\% significant} &  \textbf{$\Delta^{(test)}_{95}$}  & \textbf{$\Delta^{(test)}_{Max}$}   \\ \hline
ACE 2005 - Entities & 0.65 & 28.96\% &  1.21 & 2.53 \\ \hline
ACE 2005 - Events & 1.97 & 34.48\% &  4.32 & 9.04 \\ \hline
CoNLL 2000 - Chunking & 0.20 & 18.36\% &  0.30 & 0.56 \\ \hline
CoNLL 2003 - NER-En & 0.42 & 31.02\% &  0.83 & 1.69 \\ \hline
CoNLL 2003 - NER-De & 0.78 & 33.20\% &  1.61 & 3.36 \\ \hline
GermEval 2014 - NER-De & 0.60 & 26.80\% &  1.12 & 2.38 \\ \hline
TempEval 3 - Events & 1.19 & 10.72\% &  1.48 & 2.99 \\ \hline
\end{tabular}
\caption{The same BiLSTM-CRF approach was evaluated twice under Evaluation 1. The threshold column depicts the average difference in percentage points $F_1$-score for statistical significance with $0.04 < p <0.05$. The \textit{\% significant} column depicts the ratio how often the difference between  $A_i^{(j)}$ and $\tilde A_i^{(j)}$ is significant. $\Delta_{95}$ depicts the 95\% percentile of differences between $A_i^{(j)}$ and $\tilde A_i^{(j)}$.  $\Delta^{(test)}_{Max}$ shows the largest difference.}
\label{table_1_out_of_1}
\end{table*}

We compute in how many cases the bootstrap method finds a statistically significant difference. Further, we compute the average $F_1$ test-score difference  $\tau$ for pairs with an estimated $p$-value between 0.04 and 0.05. This value can be seen as a threshold: If the $F_1$-score difference is larger than this threshold, there is a high chance that the bootstrap method testifies a statistical significance between the two models.

Further, we compute the differences between the test performances for approach $A$ and $\tilde A$. For Evaluation 1, we compute $\Delta^{(test),(i,j)} = |\Psi^{(test)}_{A_i^{(j)}} - \Psi^{(test)}_{\tilde A_i^{(j)}}|$. For Evaluation 2, we compute:  $$\Delta^{(test),(j)} = |\Psi^{(test)}_{A_*^{(j)}} - \Psi^{(test)}_{\tilde A_*^{(j)}}|.$$ 
For those delta values we compute a 95\% percentile $\Delta^{(test)}_{95}$. The value indicates that a difference in the test score for a given task should be higher than $\Delta^{(test)}_{95}$, otherwise there is a chance greater $5\%$ that the difference is due to chance for the given task and the given network architecture.\footnote{Note that $\Delta^{(test)}_{95}$ depends on the used machine learning approach and the specific task.}

\subsection{Comparing Single Performance Scores} \label{sec_single_performance_scores}

\reftable{table_1_out_of_1} depicts the main results for Evaluation 1. For the \textit{ACE 2005 - Events} task, we observe in 34.48\% of the cases a significant difference between the models $A_i^{(j)}$ and $\tilde A_i^{(j)}$. For the other tasks, we observe similar results and between 10.72\% and 33.20\% of the cases are statistically significant.

The average $F_1$-score difference for statistical significance for the \textit{ACE 2005 - Events} task is $\tau=1.97$ percentage points. However, we observe that the difference between $A_i^{(j)}$ and $\tilde A_i^{(j)}$ can be as large as 9.04 percentage points $F_1$. While this is a rare outlier, we observe that the 95\% percentile $\Delta^{(test)}_{95}$ is more than twice as large as $\tau$ for this task and dataset.

We conclude that training two non-deterministic approaches a single time and comparing their test performances is insufficient if we are interested to find out which approach is superior for that task.

\subsection{Selecting the Best out of $n$ Runs} \label{sec_1_out_of_n}

\begin{table*}[t]
%\small
\centering
\begin{tabular}{|c|c||c|c||c|c|c|}
\hline
\textbf{Task} & \textbf{Spearman $\rho$} & \textbf{Threshold $\tau$} &\textbf{\% significant} & \textbf{$\Delta^{(dev)}_{95}$} &  \textbf{$\Delta^{(test)}_{95}$}  & \textbf{$\Delta^{(test)}_{Max}$}   \\ \hline
ACE 2005 - Entities & 0.153 & 0.65 & 24.86\% &  0.42 & 1.04 & 1.66 \\ \hline
ACE 2005 - Events & 0.241 & 1.97 & 29.08\% &  1.29 & 3.73 & 7.98 \\ \hline
CoNLL 2000 - Chunking & 0.262 & 0.20 & 15.84\% &  0.10 & 0.29 & 0.49 \\ \hline
CoNLL 2003 - NER-En & 0.234 & 0.42 & 21.72\% &  0.27 & 0.67 & 1.12 \\ \hline
CoNLL 2003 - NER-De & 0.422 & 0.78 & 25.68\% &  0.58 & 1.44 & 2.22 \\ \hline
GermEval 2014 - NER-De & 0.333 & 0.60 & 16.72\% &  0.48 & 0.90 & 1.63 \\ \hline
TempEval 3 - Events & -0.017 & 1.19 & 9.38\% &  0.74 & 1.41 & 2.57 \\ \hline
\end{tabular}
\caption{The same BiLSTM-CRF approach was evaluated twice under Evaluation 2. The threshold column depicts the average difference in percentage points $F_1$-score for statistical significance with $0.04 < p <0.05$. The \textit{\% significant} column depicts the ratio how often the difference between  $A_*^{(j)}$ and $\tilde A_*^{(j)}$ is significant. $\Delta_{95}$ depicts the 95\% percentile of differences between $A_*^{(j)}$ and $\tilde A_*^{(j)}$.  $\Delta^{(test)}_{Max}$ shows the largest difference.}
\label{table_1_out_of_n}
\end{table*}

Non-deterministic approaches can produce weak as well as strong models as shown in the previous section. Instead of training those a single time, we tune the approach and only compare the ``best'' model for each approach, i.e., the models that performed best on the development set. This evaluation method was formalized in Evaluation 2.

\reftable{table_1_out_of_n} depicts the results of this experiment. For all tasks, we observe small Spearman's rank correlation $\rho$ between the development and the test score. The low correlation indicates that a run with high development score doesn't have to yield a high test score.

For the \textit{ACE 2005 - Events} task, we observe a significant difference between $A_*^{(j)}$ and $\tilde A_*^{(j)}$ in 29.08\% of the cases. We observe for this task that the difference in test score can be as large as 7.98 percentage points $F_1$-score between $A_*^{(j)}$ and $\tilde A_*^{(j)}$.

As before, we observe that $\Delta^{(test)}_{95}$ is much larger than $\tau$, i.e.\ test performances of $A_*$ vary to a large degree, larger than the threshold $\tau$ for statistical significance.

The table also depicts $\Delta^{(dev)}_{95}$, the 95\% percentile of differences in terms of development performance. We observe a large discrepancy between $\Delta^{(dev)}_{95}$ and $\Delta^{(test)}_{95}$: For the 1,000 rows, we were able to find models $A_*^{(j)}$ and $\tilde A_*^{(j)}$ that performed comparably on the development set. However, their performance differs largely on the actual test set.

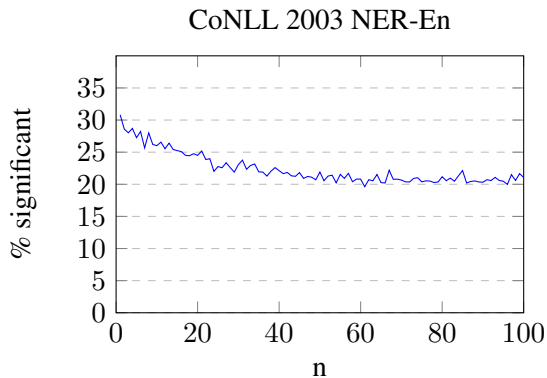
\begin{figure}[h]
\centering
\begin{tikzpicture}
\begin{axis}[
    title={CoNLL 2003 NER-En},
    xlabel={n},
    height=5cm, width=7cm,
    ylabel={\% significant},
    xmin=0, xmax=100,
    ymin=0, ymax=40,
    xtick={0,20,40,60,80,100},
    ytick={0,5,10,15,20,25,30,35},
    ymajorgrids=true,
    grid style=dashed,
]
\addplot[color=blue]
    coordinates {
(1, 30.82) (2, 28.60) (3, 28.02) (4, 28.70) (5, 27.26) (6, 28.22) (7, 25.66) (8, 28.00) (9, 26.20) (10, 26.02) (11, 26.56) (12, 25.54) (13, 26.40) (14, 25.42) (15, 25.24) (16, 25.10) (17, 24.52) (18, 24.44) (19, 24.76) (20, 24.50) (21, 25.18) (22, 23.86) (23, 23.98) (24, 22.02) (25, 22.76) (26, 22.58) (27, 23.36) (28, 22.60) (29, 21.90) (30, 23.06) (31, 23.76) (32, 22.34) (33, 22.94) (34, 23.16) (35, 21.94) (36, 21.90) (37, 21.28) (38, 22.04) (39, 22.60) (40, 22.12) (41, 21.66) (42, 21.84) (43, 21.32) (44, 21.24) (45, 21.82) (46, 20.92) (47, 21.22) (48, 21.10) (49, 20.70) (50, 21.90) (51, 20.56) (52, 21.28) (53, 21.42) (54, 20.22) (55, 21.54) (56, 20.92) (57, 21.70) (58, 20.40) (59, 20.82) (60, 20.80) (61, 19.62) (62, 20.72) (63, 20.54) (64, 21.52) (65, 20.26) (66, 20.20) (67, 22.18) (68, 20.78) (69, 20.80) (70, 20.66) (71, 20.38) (72, 20.36) (73, 20.92) (74, 21.02) (75, 20.38) (76, 20.54) (77, 20.50) (78, 20.28) (79, 20.36) (80, 21.18) (81, 20.56) (82, 20.96) (83, 20.48) (84, 21.32) (85, 22.14) (86, 20.20) (87, 20.44) (88, 20.52) (89, 20.36) (90, 20.32) (91, 20.72) (92, 20.58) (93, 21.08) (94, 20.60) (95, 20.48) (96, 19.98) (97, 21.50) (98, 20.60) (99, 21.66) (100, 21.06)
    };
 \end{axis}
\end{tikzpicture}
\caption{\small Ratio of statistically significant differences between $A_*$ and $\tilde A_*$  for different $n$-values.}
\label{fig_1_out_of_n_significant}
\end{figure}

We studied if the value of statistically significant differences between $A_*$ and $\tilde A_*$ depends on $n$, the number of sampled models. \reffigure{fig_1_out_of_n_significant} depicts the ratio for different $n$-values for the CoNLL 2003 NER task in English. We observe that the ratio of significant differences decreases with increasing number of sampled models $n$. However, the ratio stays flat after about 40 to 50 sampled models. For $n=100$ we observe that $21.06\%$ of the pairs are significant different with a $p<0.05$ value.

\section{Why Comparing Best Model Performances is Insufficient}\label{sec_proof}
While it is straightforward to understand why Evaluation 1 is improper for non-deterministic machine learning approaches, it is less obvious why this is also the case for Evaluation 2. If we ignore the bad models, where the approach did not converge to a good performance, why can't we evaluate the best achievable performances of approaches? 

The issue is not the significance test but has to do with the wrong conclusions we draw from a significant difference. The null-hypothesis for, e.g., the bootstrap test is that two compared models would perform not differently on the complete data distribution. However, it is wrong to conclude from this that one approach is capable of producing better models than the other approach. The issue is that selecting a model with high test / true performance is only possible to a certain degree and the uncertainty depends on the development set.

We write the (hypothetical) performance on the complete data distribution as $\Psi^{(true)}$. The development and test score are finite approximations of this true performance of a model.

We can rewrite the development score as $\Psi^{(dev)} = \Psi^{(true)} + \mathcal{X}^{(dev)}$ and the test score as $\Psi^{(test)} = \Psi^{(true)} + \mathcal{X}^{(test)}$. $\mathcal{X}^{(dev)}$ and $\mathcal{X}^{(test)}$ are two random variables with unknown means and variances stemming from the finite sizes of development and test set.

Given two models $A_*$ and $B_*$, the significance test checks the null hypothesis whether $\Psi^{(true)}_{A_*}$  is equal to $\Psi^{(true)}_{B_*}$ given the two results on the test set.

When we select the models $A_*$ and $B_*$ based on their performance on the development set, we face the issue that the true performance is not monotone in the development score. 

Assume we have models $A_1$ and $A_2$ with identical development performance. The development performance might be:
\begin{align*}
 \Psi^{(dev)}_{A_1} &= \Psi^{(true)}_{A_1} + \mathcal{X}^{(dev)}_{A_1} = 80\%-2\% = 78\% \\
 \Psi^{(dev)}_{A_2} &= \Psi^{(true)}_{A_2} + \mathcal{X}^{(dev)}_{A_2} = 76\%+2\% = 78\%
\end{align*}

The test performances might be:
\begin{align*}
 \Psi^{(test)}_{A_1} &= 80\%+1\% = 81\% \\
 \Psi^{(test)}_{A_2} &= 76\%-1\% = 75\%
\end{align*}

We compare this against model $B_*$ from approach $B$, which as a test performance of $ \Psi^{(test)}_{B_*} = 79\%$:

If we select $A_1$ for the comparison against $B_*$, the significance test might correctly identify that $A_1$ has a significantly lower test performance than $B_*$. However, if we select model $A_2$, the significance test might identify that $B_*$ has a significantly higher test performance than $A_2$. As we do not know which model, $A_1$ or $A_2$, to select for Evaluation 2, the outcome of Evaluation 2 is up to chance. If we select $A_1$, we might conclude that approach $A$ is better than approach $B$,  if we select $A_2$, we might conclude the opposite.

In summary, a significance test based on a single model performance can only identify which model is better but does not allow the conclusion which learning approach is superior.

\subsection{Distribution of $\Psi^{(test)}_{A_1} - \Psi^{(test)}_{A_2} $}
We are interested to which degree the test score can vary for two models with identical development scores.

We can write the scores as:
\begin{align*}
 \Psi^{(dev)}_{A_1} &= \Psi^{(true)}_{A_1} + \mathcal{X}^{(dev)}_{A_1} \\
 \Psi^{(dev)}_{A_2} &= \Psi^{(true)}_{A_2} + \mathcal{X}^{(dev)}_{A_2} \\
 \Psi^{(test)}_{A_1} &= \Psi^{(true)}_{A_1} + \mathcal{X}^{(test)}_{A_1} \\ 
 \Psi^{(test)}_{A_2} &= \Psi^{(true)}_{A_2} + \mathcal{X}^{(test)}_{A_2} 
\end{align*}

We assume $\Psi^{(dev)}_{A_1} = \Psi^{(dev)}_{A_2}$, hence:
 \begin{gather*}
  \Psi^{(true)}_{A_1} + \mathcal{X}^{(dev)}_{A_1} = \Psi^{(true)}_{A_2} + \mathcal{X}^{(dev)}_{A_2} \\
  \Rightarrow \Psi^{(true)}_{A_1} - \Psi^{(true)}_{A_2} = \mathcal{X}^{(dev)}_{A_2} - \mathcal{X}^{(dev)}_{A_1} 
 \end{gather*}

For the test performance difference, this leads to: 
 \begin{gather*}
  \Psi^{(test)}_{A_1} - \Psi^{(test)}_{A_2} \\
  = (\Psi^{(true)}_{A_1} - \Psi^{(true)}_{A_2}) + (\mathcal{X}^{(test)}_{A_1} - \mathcal{X}^{(test)}_{A_2}) \\
  = (\mathcal{X}^{(dev)}_{A_2} - \mathcal{X}^{(dev)}_{A_1}) + (\mathcal{X}^{(test)}_{A_1} - \mathcal{X}^{(test)}_{A_2})
\end{gather*}

The difference in test performance between $A_1$ and $A_2$ does not only depend on $\mathcal{X}^{(test)}$, but also on the random variable of the development set  $\mathcal{X}^{(dev)}$. Hence, the variance introduced by the finite approximation of the development set is important to understand the variance of test scores.

\subsection{Emperical Estimation}

In this section we study how large the test score can vary for the studied tasks from \refsec{sec_experiment}. We assume $\Psi^{(dev)}_{A_1} = \Psi^{(dev)}_{A_2}$. We are interested in how much the test score for these two models can vary, i.e.\ how large the difference $|\Psi^{(test)}_{A_1} - \Psi^{(test)}_{A_2}|$ can reasonably become.

We do this by computing a  linear regression $f(\Psi^{(dev)}) \approx \Psi^{(test)}$ between the development and test score. For this linear regression, we compute the prediction interval $\zeta$ \cite{Faraway2002}. The test score should be within the range $f(\Psi^{(dev)}) \pm \zeta(\Psi^{(dev)})$ with a confidence of $\alpha$.

The prediction interval is given by:
$$\zeta(\Psi^{(dev)}) = t^*_{n-2} s_y \sqrt{1 + \frac{1}{n} + \frac{(\Psi^{(dev)} - \overline{\Psi^{(dev)}}) ^2}{(n-1)s_x^2}}$$

with $n$ the number of samples, $t^*_{n-2}$ the value for the two-tailed t-distribution at the desired confidence $\alpha$ for the value $n-2$, $s_y$ the standard deviation of the residuals calculated as:
$$s_y = \sqrt{\frac{\sum(\Psi^{(test)} - \hat\Psi^{(test)})^2}{n-2}}$$

$\overline{\Psi^{(dev)}}$ the mean value $\Psi^{(dev)}_i$ and $s_x$ the unbiased estimation of standard deviation:
$$s_x^2 = \frac{1}{n-1} \sum_{i=1}^n \left(\Psi^{(dev)}_i - \overline{\Psi^{(dev)}}\right)^2.$$

An extreme difference in test score would be $\Psi^{(test)}_{A_1} \leq f(\Psi^{(dev)}) - \zeta(\Psi^{(dev)})$ for the one model and $\Psi^{(test)}_{A_2}  \geq f(\Psi^{(dev)}) + \zeta(\Psi^{(dev)})$ for the other model. The difference would then be $|\Psi^{(test)}_{A_1} - \Psi^{(test)}_{A_2}| \geq 2\zeta(\Psi^{(dev)})$.

The probability of $|\Psi^{(test)}_{A_1} - \Psi^{(test)}_{A_2}| \geq 2\zeta(\Psi^{(dev)})$ is  $(1-\alpha)^2$. We set $(1-\alpha)^2 = 0.05$. In this case, $|\Psi^{(test)}_{A_1} - \Psi^{(test)}_{A_2}| \leq 2\zeta(\Psi^{(dev)})$ in 95\% of the cases.

The value of $2\zeta(\Psi^{(dev)})$ is approximately constant in terms of the development score $\Psi^{(dev)}$. Hence, we computed the mean $2\overline{\zeta\Psi^{(dev)})}$ and depict the value in \reftable{table_prediction_interval}. 

\begin{table}[ht]
%\small
\centering
\begin{tabular}{|c|c|}
\hline
\textbf{Task} & \textbf{Predict. Interval}  \\ \hline
ACE 2005 - Entities & $1.03$ \\ \hline
ACE 2005 - Events & $3.68$ \\ \hline
CoNLL 2000 - Chunking & $0.25$ \\ \hline
CoNLL 2003 - NER-En & $0.69$ \\ \hline
CoNLL 2003 - NER-De & $1.24$ \\ \hline
GermEval 2014 - NER-De & $0.88$ \\ \hline
TempEval 3 - Events & $1.30$ \\ \hline
\end{tabular}
\caption{\small Size of the 95\% interval for the test scores of two models with the same development score. }
\label{table_prediction_interval}
\end{table}

The value $3.68$ for the \textit{ACE 2005 - Events} tasks indicates that, given two models with the same performance on the development set, the test performance can vary up to 3.68 percentage points $F_1$-score (95\% interval). The values $2\overline{\zeta(\Psi^{(dev)})}$ are comparably similar to the value of $\Delta_{95}^{(test)}$ in \reftable{table_1_out_of_n}.

\section{Evaluation Methodologies based on Score Distributions} \label{sec_definition_score_distributions}
In this section, we formally define two  idealized definitions for \textit{approach A superior to approach B}.

We define the performance for a model as:
\begin{align}
\Psi^{(Test)}_{A_{(\mathtt{Train}, \mathtt{Dev}, \mathtt{Rnd})}} = S(A_{(\mathtt{Train}, \mathtt{Dev}, \mathtt{Rnd})}(\mathtt{Test}_x), \mathtt{Test}_y).
\end{align}

$A$ is the learning approach that trains a model given a training set \texttt{Train}, a development set \texttt{Dev} and a sequence of random numbers \texttt{Rnd}. The resulting model $A_{(\mathtt{Train}, \mathtt{Dev}, \mathtt{Rnd})}$ is applied to the test dataset \texttt{Test}$_x$ and a performance score $S$ is computed between the predictions and the gold labels \texttt{Test}$_y$.

\textbf{Evaluation 3.} Given a certain task and a potentially infinite data population $\mathcal{D}$. We call \textit{approach A superior to approach B} for this task with training set of size $k \leq |\mathtt{Train}| \leq l$ if and only if the expected test score for approach $A$ is larger than the expected test score for approach $B$:
\begin{align*}
 E\left[\Psi^{(Test)}_{A_{(\mathtt{Train}, \mathtt{Dev}, \mathtt{Rnd})}}\right] > E\left[\Psi^{(Test)}_{B_{(\mathtt{Train}, \mathtt{Dev}, \mathtt{Rnd})}}\right] 
\end{align*}
with \texttt{Train}, \texttt{Dev}, and \texttt{Test} sampled from $\mathcal{D}$.

We can approximate the expected test score for an approach by training multiple models and comparing the sample mean values $\overline{\Psi^{(Test)}_{A_{1...n}}}$  and $\overline{\Psi^{(Test)}_{B_{1...m}}}$. We conclude that one approach is superior if the difference between the means is significant. 

A common significance test used in literature is the Welch's t-test. This is a simple significance test which only requires the information on the sample mean, sample variance and sample size. However, the test assumes that the two distributions are approximately normally distributed.

Evaluation 3 computes the expected test score, however, \textit{superior} can also be interpreted as a higher probability to produce a better working model.

\textbf{Evaluation 4.} Given a certain task and a potentially infinite data population $\mathcal{D}$. We call \textit{approach A superior to approach B} for this task with training set of size $k \leq |\mathtt{Train}| \leq l$ if and only if the probability for approach $A$ is higher to produce a better working model than it is for approach $B$.  We call approach $A$ superior to approach $B$ if and only if:
\begin{align*}
 P\left( \Psi^{(Test)}_{A_{(\mathtt{Train}, \mathtt{Dev}, \mathtt{Rnd})}} \geq  \Psi^{(Test)}_{B_{(\mathtt{Train}, \mathtt{Dev}, \mathtt{Rnd})}} \right) > 0.5
\end{align*}

We can estimate if the probability is significantly different from 0.5 by sampling a sufficiently large number of models from approach $A$ and approach $B$ and then applying either a  Mann-Whitney U test for independent pairs or a Wilcoxon signed-rank test for matched (dependent) pairs for the achieved test scores.

In contrast to the Welch's t-test, those two tests do not assume a normal distribution. To perform the Wilcoxon signed-rank test, at least 6 models for a two-tailed test are needed to be able to get a confidence level $p < 0.05$ \cite{Sani2005}. For a confidence level of $p < 0.01$, at least 8 models are needed. 

There is a fine distinction between Evaluation 3 and Evaluation 4. Evaluation 3 compares the mean values for two approaches, while Evaluation 4 compares the medians of the distributions.\footnote{Note, for certain distributions, the median $m$ with $P(X \leq m) \leq 0.5$ and $P(X \geq m) \leq 0.5$ might not be uniquely defined. This does not affect Evaluation 4.} For skewed distributions, the median is different from the mean, which might change the drawn conclusion from Evaluation 3 and Evaluation 4. Approach $A$ might have a better mean score than approach $B$, but a lower median than approach $B$ or vice versa. 

Note, \texttt{Train}, \texttt{Dev}, and \texttt{Test} in Evaluation 3 and 4 are random variables sampled from the (infinite) data population $\mathcal{D}$. This is an idealized formulation for comparing machine learning approaches as it assumes that new, independent datasets from $\mathcal{D}$ can be sampled. However, for most tasks, it is not easily possible to sample new datasets. Instead, only a finite dataset is labeled that must be used for \texttt{Train}, \texttt{Dev}, and \texttt{Test}. This creates the risk that an approach might be superior for a specific dataset, however, for other train, development, or test sets, this might not be the case. In contrast, addressing the variation introduced by $\mathtt{Rnd}$ is straightforward by training the approach with multiple random sequences. 

Evaluation 3 and Evaluation 4 both mention that training sets are of size $k \leq |\mathtt{Train}| \leq l$. Learning approaches can react differently to increasing or decreasing training set sizes, e.g., approach $A$ might be better for larger training sets while approach $B$ might be better for smaller training sets. When comparing approaches, it would be of interest to know the lower bound $k$ and the upper bound $l$ for approaches $A$ and $B$. However, most evaluations check for practical reasons only one training set size, i.e., $k=l$.

\section{Experiment (Score Distributions)} \label{sec_results_score_distributions}
In this section, we study if Evaluation 3 and Evaluation 4 can reliably detect that there is no difference between approach $A$ and $\tilde A$ from \refsec{sec_experiment}.

We compare 25 models from approach $A$ ($A_1^{(j)}, ..., A_{25}^{(j)}$) with 25 models from approach $\tilde A$ ($\tilde A_1^{(j)}, ..., \tilde A_{25}^{(j)}$) each trained with a different random sequence $\mathtt{Rnd}$. For Evaluation 3, we use Welch's t-test, for Evaluation 4, Wilcoxon signed-rank test. As threshold, we used $p<0.05$.

\begin{table}[ht]
%\small
\centering
\begin{tabular}{|c|c|c|}
\hline
\textbf{Task} & \textbf{Eval. 3} & \textbf{Eval. 4}    \\ \hline
ACE - Entities & 4.68\% & 4.86\%  \\ \hline
ACE - Events & 4.72\% & 4.67\%  \\ \hline
CoNLL - Chunking & 4.60\% & 4.86\%  \\ \hline
CoNLL - NER-En & 5.18\% & 5.01\%   \\ \hline
CoNLL - NER-De & 4.83\% & 4.78\%  \\ \hline
GermEval - NER-De & 4.91\% & 4.74\%   \\ \hline
TempEval - Events & 4.72\% & 5.03\%  \\ \hline
\end{tabular}
\caption{\small Percentage of significant difference between $A$ and $\tilde A$ for $p<0.05$. }
\label{table_score_dist_significant}
\end{table}

\reftable{table_score_dist_significant} summarizes the outcome of this experiment. The ratios are all at about 5\%, which is the number of false positives we would expect from a threshold $p<0.05$. In contrast to Evaluation 1 and 2, Evaluation 3 and 4 were able to identify that the approaches are identical in most cases.

Next, we study how stable the mean $\overline{\Psi^{(test)}_{A_{1,...,n}^{(j)}}}$ is for various values of $n$. The larger the variance, the more difficult will it be to spot a difference between two learning approaches. To express the variance in an intuitive value, we compute the 95th percentile $\Delta_{95}^{(test)}$ for the difference between the mean scores:
$$\Delta^{(test),(n,j)} = \left | \overline{\Psi^{(test)}_{A_{1,...,n}^{(j)}}} - \overline{\Psi^{(test)}_{\tilde A_{1,...,n}^{(j)}}} \right |$$ 
The value $\Delta_{95}^{(test)}$ gives an impression which improvement in mean test score is needed for a significant difference. Note, this value depends on the variance of the produced models.

The values are depicted in \reftable{table_mean_delta95}. For increasing $n$ the value $\Delta_{95}^{(test)}$ decreases, i.e.\ the mean score becomes more stable. However, for the CoNLL 2003 NER-En task we still observe a difference of 0.26 percentage points $F_1$-score between the mean scores for $n=10$. For the ACE 2005 Events dataset, the value is even at $1.39$ percentage points $F_1$-score.

\begin{table}[ht]
%\small
\centering
\begin{tabular}{|c|c|c|c|c|c|}
\hline
& \multicolumn{5}{c|}{\textbf{$\Delta_{95}^{(test)}$ for $n$ scores}}   \\ \hline
\textbf{Task}  & \textbf{1} & \textbf{3} & \textbf{5} & \textbf{10} & \textbf{20} \\ \hline
ACE-Ent. & 1.21 & 0.72 & 0.51 & 0.38 & 0.26 \\ \hline
ACE-Ev. & 4.32 & 2.41 & 1.93 & 1.39 & 0.97 \\ \hline
Chk. & 0.30 & 0.16 & 0.14 & 0.09 & 0.06 \\ \hline
NER-En & 0.83 & 0.45 & 0.35 & 0.26 & 0.18 \\ \hline
NER-De & 1.61 & 0.94 & 0.72 & 0.51 & 0.37 \\ \hline
GE 14 & 1.12 & 0.64 & 0.48 & 0.34 & 0.25 \\ \hline
TE 3 & 1.48 & 0.81 & 0.63 & 0.48 & 0.32 \\ \hline
\end{tabular}
\caption{\small 95\% percentile of $\Delta^{(test)}$ after averaging. }
\label{table_mean_delta95}
\end{table}

\section{Discussion \& Conclusion} \label{sec_conclusion}

Non-deterministic approaches like neural networks can produce models with varying performances and comparing performances based on single models does not allow drawing conclusions about the underlying learning approaches.

An interesting observation is that the variance of the test scores depends on the development set. With an improper development set, the achieved test scores for the same approach can vary arbitrarily large. Without a good development set, we face the challenge of not knowing which configuration in weight space to choose.

We conclude that the meaningfulness of a test score is limited by the quality of the development set. This is an important observation, as often little attention is paid to the selection of the development set. To have as much training data as possible, we often prefer small development sets, sometimes substantially smaller than the test set. 

Future work is needed to judge the importance of the development set and how to select it appropriately. As of now, we recommend using a development set that is of comparable size to the test set.

For the organization of shared tasks, we recommend that participants do not submit only a single model, but multiple models trained with different random seed values. Those submissions should not be treated individually. Instead the mean and the standard deviation of test scores should be reported.

Previous work showed that there can be large differences between local minima of neural networks and that some minima generalize badly to unseen data. Those minima also generalize badly on the development set and do not play a role in the final evaluation. This form of evaluation, where only the model that performed best on the development set is evaluated on unseen test data, can be found in many publications and many shared tasks evaluate individual models submitted by the participants.

We showed that this evaluation setup is not suitable to draw conclusions about machine learning approaches. A statistically significant difference of test scores does not have to be the result of a superior learning approach. There is a high risk that this is due to chance. Further, we showed that the development set has a major impact on the test score variance. 

Our observations are not limited to non-deterministic machine learning approaches. If we treat hyperparameters as part of an approach, it also affects deterministic approaches like support vector machines. For an SVM we might achieve with two slightly different configurations identical development scores, however, both models might show a large difference in terms of test score. It is up to chance which model would be select for the final evaluation.

We provide two formalizations for comparing learning approaches. The first compares expected scores, however, it requires that scores are approximately normal distributed for significance testing. The second defines superiority of a learning approach in terms of the probability to produce a better working model. This definition can be tested without the assumption of normal distributed scores. For the evaluated approach and tasks, we showed that the type I error rate matches the $p$-value of the significance tests.

For shared tasks, we propose that participants submit multiple models, at least 6 for a $p$-value of $0.05$, trained with different sequences of random numbers.  Those submissions should not be treated individually. Instead we recommend the comparison of score distributions.

\FloatBarrier
% include your own bib file like this:
\bibliography{references}
\bibliographystyle{acl2012}

\end{document}